\documentclass[conference]{IEEEtran}
\IEEEoverridecommandlockouts
\usepackage{cite}
\usepackage{amsmath,amssymb,amsfonts}
\usepackage{algorithmic}
\usepackage[draft]{hyperref}
\usepackage{graphicx}
\usepackage{textcomp}
\usepackage{xcolor}
\usepackage[]{algorithm2e}
\usepackage{footnote}
\usepackage{caption}
\usepackage{subcaption}

\begin{document}


\author
{\IEEEauthorblockN{Divyam Madaan$^{*}$\thanks{*Equal Contribution}}
\IEEEauthorblockA{
\textit{Panjab University}\\
divyam3897@gmail.com}
\and
\IEEEauthorblockN{Radhika Dua$^{*}$}
\IEEEauthorblockA{
\textit{Panjab University}\\
radhikadua1997@gmail.com}
\and
\IEEEauthorblockN{Prerana Mukherjee}
\IEEEauthorblockA{
\textit{IIIT Sricity}\\
prerana.m@iiits.in}
\and
\IEEEauthorblockN{Brejesh Lall}
\IEEEauthorblockA{
\textit{IIT Delhi}\\
brejesh@ee.iitd.ac.in}
}

\title{VayuAnukulani: Adaptive Memory Networks for Air Pollution Forecasting}

\maketitle

\begin{abstract}
Air pollution is the leading environmental health hazard globally due to various sources which include factory emissions, car exhaust and cooking stoves. As a precautionary measure, air pollution forecast serves as the basis for taking effective pollution control measures, and accurate air pollution forecasting has become an important task. In this paper, we forecast fine-grained ambient air quality information for 5 prominent locations in Delhi based on the historical and real-time ambient air quality and meteorological data reported by Central Pollution Control board. We present VayuAnukulani system, a novel end-to-end solution to predict air quality for next 24 hours by estimating the concentration and level of different air pollutants including nitrogen dioxide ($NO_2$), particulate matter ($PM_{2.5}$ and $PM_{10}$) for Delhi. Extensive experiments on data sources obtained in Delhi demonstrate that the proposed adaptive attention based Bidirectional LSTM Network outperforms several baselines for classification and regression models. The accuracy of the proposed adaptive system is $\sim 15 - 20\%$ better than the same offline trained model.
We compare the proposed methodology on several competing baselines, and show that the network outperforms conventional methods by $\sim 3 - 5 \%$.
\end{abstract}


\section{INTRODUCTION}
Due to the rapid urbanization and industrialization, there is an increase in the number of vehicles, and the burning of fossil fuels, due to which the quality of air is degrading, which is a basic requirement for the survival of all lives on Earth. The rise in the pollution rate has affected people with serious health hazards such as heart disease, lung cancer and chronic respiratory infections like asthma, pneumonia etc. Air pollution has also become a major cause for skin and eye health conditions in the country.
 
One of the most important emerging environmental issues in the world is air pollution. According to World Health Organization (WHO), in 2018, India has 14 out of the 15 most polluted cities in the world. The increasing level of pollutants in ambient air in 2016-2018 has deteriorated the air quality of Delhi at an alarming rate. All these problems brought us to focus the current study on air quality in Delhi region. The establishment of a reasonable and accurate forecasting model is the basis for forecasting urban air pollution, which can inform government's policy-making bodies to perform traffic control when the air is polluted at critical levels and people's decision making like whether to exercise outdoors or which route to follow.

\begin{figure}[h!]
    \centering
    \includegraphics[width=8cm, height=5cm]{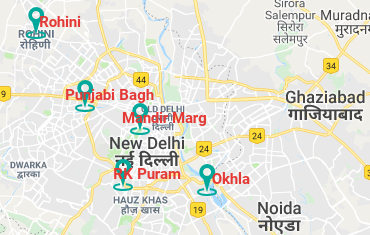}
    \caption{{Map showing air quality monitoring stations of Delhi}}%
    \label{fig:location}%
\end{figure}

%
Prior work \cite{dong2009,Sun2017, zheng15} focuses on applying the regression and neural network  models for forecasting  $PM_{2.5}$ concentration based on weather and air quality data . However, these works do not take into consideration the crucial aspect of predicting the concentration and level of other pollutants which are equally important and have serious impact on health conditions. Usually, the concentration of  $PM_{2.5}$ is in the normal range but the concentration of other pollutants is very high. Therefore, in this paper, we aim towards addressing this gap and utilize a novel distributed architecture to simultaneously predict multiple pollutants present in the ambient air.

Also, the current air pollution forecasting models have been studied over existing datasets with the objective to maximize the accuracy. However, continuous online training means the accuracy of the model trained offline varies. We study the challenge of estimating the parameters of the new model by choosing the highest accuracy model from an ensemble of deep learning and classical machine learning models.

The system we present, VayuAnukulani\footnote{The name is a direct Hindi translation for the two key words Air and Adaptatability - Air, translated as Vayu and Adaptatability,
translated as Anukulani.
}, predicts the air quality by estimating the concentration and  level of  different  air  pollutants and can  pre-inform government’s  policy-making  bodies  to  undertake preventive measures.
In this paper, we 
propose a novel approach to predict the concentration and pollution level i.e low, moderate and high pollution for different air pollutants including nitrogen dioxide ($NO_2$) and particulate matter ($PM_{2.5}$ and $PM_{10}$) for 5 different monitoring stations in Delhi as shown in Fig. \ref{fig:location} (screenshot from our website), considering air quality data, time and meteorological data of the past 24 hours. To the best of the author's knowledge, this work is one of the first holistic and empirical research where a single model predicts multiple pollutants and makes adaptive updates to improvise the future predictions.

The key contributions in this work can be summarized as,
\begin{enumerate}

\item The proposed approach uses heterogeneous urban data for capturing both direct (air pollutants) and indirect (meteorological data and time) factors. The model adapts to a novel distributed architecture in order to simultaneously model the interactions between these factors for leveraging learning of the individual and holistic influences on air pollution.

\item We build a novel end-to-end adaptive system that collects the pollution data over a period of 1-3 years for 5 locations of Delhi, trains two models (forecast/classify) to predict the concentration and classify the pollution level for each pollutant  to 2-3 classes (3 classes for $PM_{2.5}$ and $PM_{10}$, 2 classes for  $SO_2$ and $NO_2$) for the next 24 hours, and then deploys a new model.

\item We achieve a significant gain of $\sim 15 - 20\%$  accuracy with the proposed adaptive system as compared to the same offline trained model. The proposed system also outperforms several competing baselines and we achieve a performance boost of about $\sim 3 - 5 \%$ in forecasting of pollution levels and concentration.

The remainder of this paper is organized as follows. In Sec. \ref{sec:challenges}, we discuss the challenges occurring in implementing air pollution forecasting system and in Sec. \ref{sec:background}, we outline the background description of various deep learning architectures for temporal forecasting. In Sec. \ref{sec:proposed}, we provide a detailed overview of the component blocks of the proposed architecture for air pollution forecasting followed by an evaluation of the same in Sec. \ref{sec:evaluation}. In Sec. \ref{sec:webapp} we provide the details about the developed web application and in Sec. \ref{sec:conclusion}, we conclude the paper.

\end{enumerate}

\section{{CHALLENGES}}
\label{sec:challenges}
 Some of the major problems that occur while implementing a prediction system to forecast the concentration and level of air pollutants are:
\begin{itemize} 
\item 
 Air pollution varies with time and space. 
Therefore, it is essential to maintain spatial and temporal granularity in air quality forecasting system.
\item The data is often insufficient and inaccurate. Sometimes the concentration of pollutants is much beyond the permissible limits.
\item Air pollution dataset contains several outliers (the pollution increases/decreases significantly all of a sudden), which is mainly due to accidental forest fires,  burning of agricultural wastes and burning of crackers during diwali.
\item Air pollution depends on direct (historical air pollution data) and indirect factors (like time, meteorological data) and modelling the relationship between these factors is quite complex.
\item In order to forecast the concentration and level of air pollutants, it is necessary to collect data for every location independently as pollution varies drastically with geographical distance.
\item Separate models are deployed for each location and training each model regularly is computationally expensive. Thus, updating data and training model for each location becomes unmanageable when we deal with many locations.
\end{itemize}

In the proposed method, we address the aforementioned challenges. We train separate models for each location as the pollution varies with varying geographical distance. Hourly air pollution and meteorological data is collected as the pollution varies significantly with time.
We use various imputation techniques in order to deal with missing values and perform data preprocessing to select and use only the important features for air pollution prediction as detailed in sec. \ref{sec:datapreprocessing}.

\section{{Related Work}}
\label{sec:related work}
In this section we provide a detailed overview of the contemporary techniques prevalent in the domains which are closely related to our work and broadly categorize them into the following categories for temporal forecasting.

\subsection{Handcrafted features based techniques}
Traditional handcrafted feature based approaches \cite{dong2009} utilize Markovian approaches to predict $PM_{2.5}$ concentration using time series data. Since, Hidden Markov Models (HMMs) suffer from the limitation of short term memory thus failing in capturing the temporal dependencies in prediction problems. To overcome this, authors in \cite{dong2009} highlight a variant of HMMs: hidden semi-Markov models (HSMMs) for $PM_{2.5}$ concentration prediction by introducing temporal structures into standard HMMs. In \cite{Sun2017}, authors provide a hybrid framework which relies on feature reduction followed by a hyperplane based classifier (Least Square Support Vector Machine)  to predict the particulate matter concentration in the ambient air. The generalization ability of the classifier is improvised using cuckoo search optimization strategy. In \cite{zhang2018prediction}, authors compare and contrast various neural network based models like Extreme Learning Machine, Wavelet Neural Networks, Fuzzy Neural Networks, Least Square Support Vector Machines to predict $PM_{2.5}$ concentration over short term durations. They demonstrate the efficacy of Wavelet Neural Networks over the compared architectures in terms of higher precision and self learning ability. Even meteorological parameters such as humidity and temperature are considered for evaluation. In \cite{shang2019novel}, authors provide an hourly prediction over $PM_{2.5}$ concentration depicting multiple change patterns. They train separate models over clusters comprising of subsets of the data utilizing an ensemble of classification and regression trees which captures the multiple change pattern in the data.

In contrast to the aligning works \cite{dong2009, zhang2018prediction, shang2019novel, Sun2017}, in this paper we capture the long-term temporal dependencies between the data sources collected using both direct as well as indirect sources and introduce the importance of attention based adaptive learning in the memory networks. Apart from that, we provide a standalone model to predict various pollutants ($PM_{2.5}, NO_2$ and $Pm_{10}$) over a long period of time in future (next 24 hours).

\subsection{Deep Learning based techniques}
In recent times, researchers are predominantly utilizing deep learning based architectures for forecasting  expressway $PM_{2.5}$ concentration. In \cite{li2017deep}, the authors devise a novel algorithm to handle the missing values in the urban data for China and Hong Kong followed by deep learning paradigm for air pollution estimation. This enables in predicting the air quality estimates throughout the city thus, curtailing the cost which might be incurred due to setup of sophisticated air quality monitors. In \cite{pengfei2018industrial}, authors provide a deep neural network (DNN) model to predict industrial air pollution. A non-linear activation function in the hidden units in the DNN reduces the vanishing gradient effect. In \cite{kok2017deep}, authors propose Long Short Term Memory (LSTM) networks to predict future values of air quality in smart cities utilizing the IoT smart city data. In \cite{li2017estimating}, authors utilize geographical distance and spatiotemporally correlated $PM_{2.5}$ concentration in a deep belief network to capture necessary evaluators associated with $PM_{2.5}$ from latent factors. In \cite{qi2018deep}, authors adopt a Deep Air Learning framework which emphasize on utilizing the information embedded in the unlabeled spatio-temporal data for  improving the interpolation and prediction performance. 

In this paper, we however predict other pollutants ($PM_{2.5}, NO_2$ and $Pm_{10}$) which are equally important to scale the air quality index to an associated harmful level utilizing the deep learning based paradigm.

\subsection{Wireless networks based techniques}
In \cite{boubrima2017poster}, authors utilize mobile wireless sensor networks (MWSN) for generating precise pollution
maps based on the sensor measurement values and physical models
which emulate the phenomenon of pollution dispersion. This works in a synergistic manner by reducing the errors due to the simulation runs based on physical models while relying on sensor measurements. However, it would still rely on the deployment of the sensor nodes. In \cite{kizel2018node}, authors utilize node-to-node calibration, in which at a time one sensor in each chain is directly calibrated against the reference measurements and the rest are calibrated sequentially while they are deployed and occur in pairs. This deployment reduces the load on sensor relocations and ensures simultaneous calibration and data collection. In \cite{feng2018air}, authors propose a framework for air quality estimation utilizing multi-source heterogeneous data collected from wireless sensor networks.

\subsection{Trends and seasonality assessment based techniques}
In \cite{mohan2007analysis}, authors present a detailed survey on the trends of the seasonal variations across Delhi regions and present analysis on the typical rise of the pollution index during certain times over the year. In \cite{peng2005seasonal}, authors investigate Bayesian semi-parametric hierarchical models for studying the time-varying effects of pollution. In \cite{chen2013seasonal}, authors study on the seasonality trends on association between particulate matter and mortality rates in China. The time-series model is utilized after smooth adjustment for time-varying confounders using natural splines. In \cite{guerreiro2014air}, authors analyze status and trends of air quality in Europe based on the ambient air measurements along with anthropogenic emission data. 

Relative to these works, we study the seasonal variations at various locations in Delhi and find the analogies for the typical rise in the pollution trends at particular locations utilizing the meteorological data and historic data.

\section{{BACKGROUND}}
\label{sec:background}
\subsection{{Long Short Term Memory Units}}
Given an input sequence $x = (x_1, ... , x_T )$, a standard recurrent neural network (RNN) computes the hidden vector sequence $h = (h_1, ... ,h_T )$ and output vector sequence $y = (y_1, ... , y_T )$ by iterating from $t = [1 ... T]$ as,
\begin{equation}
h_t = H(W_{xh} x_t + W_{hh} h_{t - 1} + b_h )    
\end{equation}
\begin{equation}
y_t = W_{hy} h_t + b_y
\end{equation}
where $W$ denotes weight matrices ($W_{xh}$ is the input-hidden weight matrix), $b$ denotes bias vectors
($b_h$ is hidden bias vector) and $H$ is the hidden layer function. $H$ is usually an element wise application of a sigmoid function given as,
\begin{equation}
f(x) = \dfrac{1}{1+e^{-x}}     
\end{equation}

It has been observed that it is difficult to train RNNs to capture long-term dependencies because the gradients tend to either vanish (most of the time) or explode (rarely, but with severe effects) \cite{rnn}. To handle this problem several sophisticated recurrent architectures like Long Short Term Memory \cite{lstm} \cite{graveslstm} and Gated Recurrent Memory \cite{gru} are precisely designed to escape the long term dependencies of recurrent networks.
LSTMs can capture long term dependencies including contextual information in the data. Since more previous information may affect the accuracy of model, LSTMs become a natural choice of use. The units can decide to overwrite the memory cell, retrieve it, or keep it for the next time step. 
The LSTM architecture can be outlined by the following equations,
\begin{equation}
   i_t = \sigma(W_{xi} x_t + W_{hi} h_{t - 1} + b_i )
   \label{eq:3}
\end{equation}
\begin{equation}
    f_t = \sigma(W_{xf} x_t + W_{hf} h_{t - 1} + b_f )
    \label{eq:4}
\end{equation}
\begin{equation}
o_t = \sigma(W_{xo} x_t + W_{ho} h_{t - 1} + b_o )
\label{eq:5}
\end{equation}
\begin{equation}
c_t = f_t ∗ c_{t−1} + i_t ∗ \tanh (W_{xg} x_t + b_g )
\label{eq:6}
\end{equation}
\begin{equation}
h_t = o_t ∗ \tanh (c_t )
\label{eq:7}
\end{equation}

where $\sigma$ is the logistic sigmoid function, and $i$, $f$, $o$ and $c$ denote the input gate, forget gate, output gate, and cell activation vectors respectively, having same dimensionality as the hidden vector $h$. The weight matrices from the cell to gate vectors denoted by $W_{xi}$  are diagonal, so element $m$ in each gate vector only receives input from element $m$ of the cell vector.

\subsection{{Bidirectional LSTM Unit}} \label{4b}
\begin{figure}
    \centering
    \includegraphics[width=8cm, height=6cm]{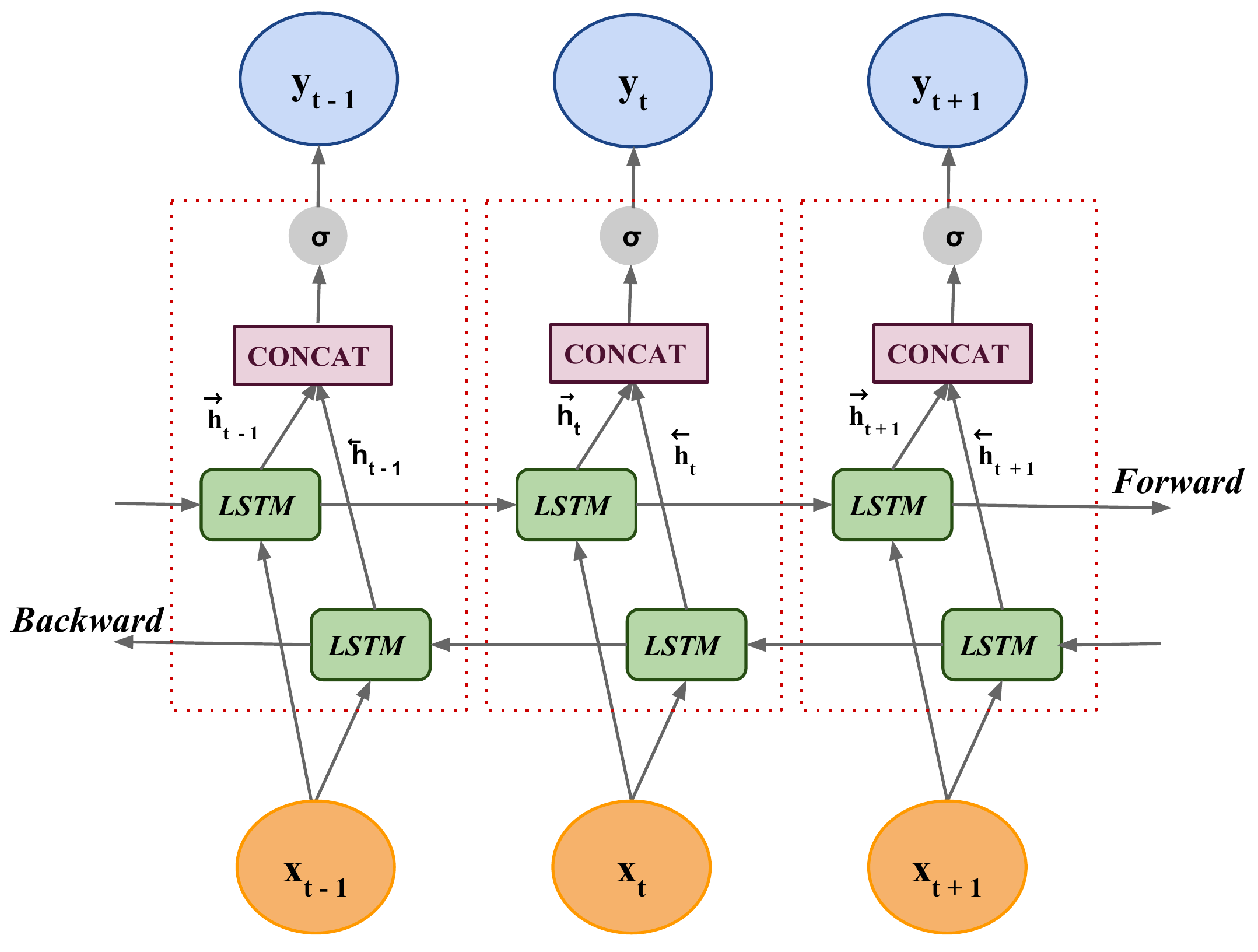}
    \caption{{Unfolded architecture of BiLSTM with 3 consecutive steps. $x_{t-1},x_t,x_{t+1}$ represent the inputs to the BiLSTM steps and $y_{t-1}, y_t, y_{t+1}$ represent the outputs of the BiLSTM steps.}}
    \label{fig:bilstm}
\end{figure}

Bidirectional LSTM (BiLSTM)  \cite{bilstm} utilizes two LSTMs to process sequence in two directions: forward and backward as shown in Fig. \ref{fig:bilstm}. The forward layer output sequence $\overrightarrow{h}$, is iteratively calculated using inputs in a positive sequence from time $T- 1$ to time $T + 1$,  while the backward layer output sequence $\overleftarrow{h}$, is calculated using the reversed inputs from time $T + 1$ to time $T - 1$. Both the forward and backward layer outputs are calculated by using the standard LSTM eqns. \ref{eq:3} - \ref{eq:8}.

BiLSTM generates  an  output  vector $Y_t$, in which each element is calculated as follows,
\begin{equation}
    y_t = \sigma(\overrightarrow{h}, \overleftarrow{h})
    \label{eq:8}
\end{equation}
where $\sigma$ function is used to combine the two output sequences. It can be a concatenating function, a summation function, an average  function or a multiplication function. Similar to the LSTM layer, the final output of a BiLSTM layer can be represented by a vector, $Y_t = [Y_{t-n},....y_{t - 1}]$  in which the last element $y_{t-1}$, is the predicted air pollution prediction for the next time iteration.

\subsection{{Attention mechanism}}
Attentive neural networks have recently demonstrated success in a wide range of tasks ranging from question answering \cite{Hermann2015TeachingMT}, machine translations \cite{attentionall} to speech recognition \cite{speechrecog}. In this section, we propose the attention mechanism for pollution forecasting tasks which assesses the importance of the representations of the encoding information $e_i$ and computes a weighted sum:
\begin{equation}
    c_j = \dfrac{1}{E} \sum_{i=1}^{E} w_{ji} e_i \\
\end{equation}
where the weights are learned through an additional attention function, which is implemented as a feed-forward neural network.

\section{{Proposed Methodology}}
\label{sec:proposed}
\begin{figure}
    \centering
    {{\includegraphics[width=9cm, height=5cm]{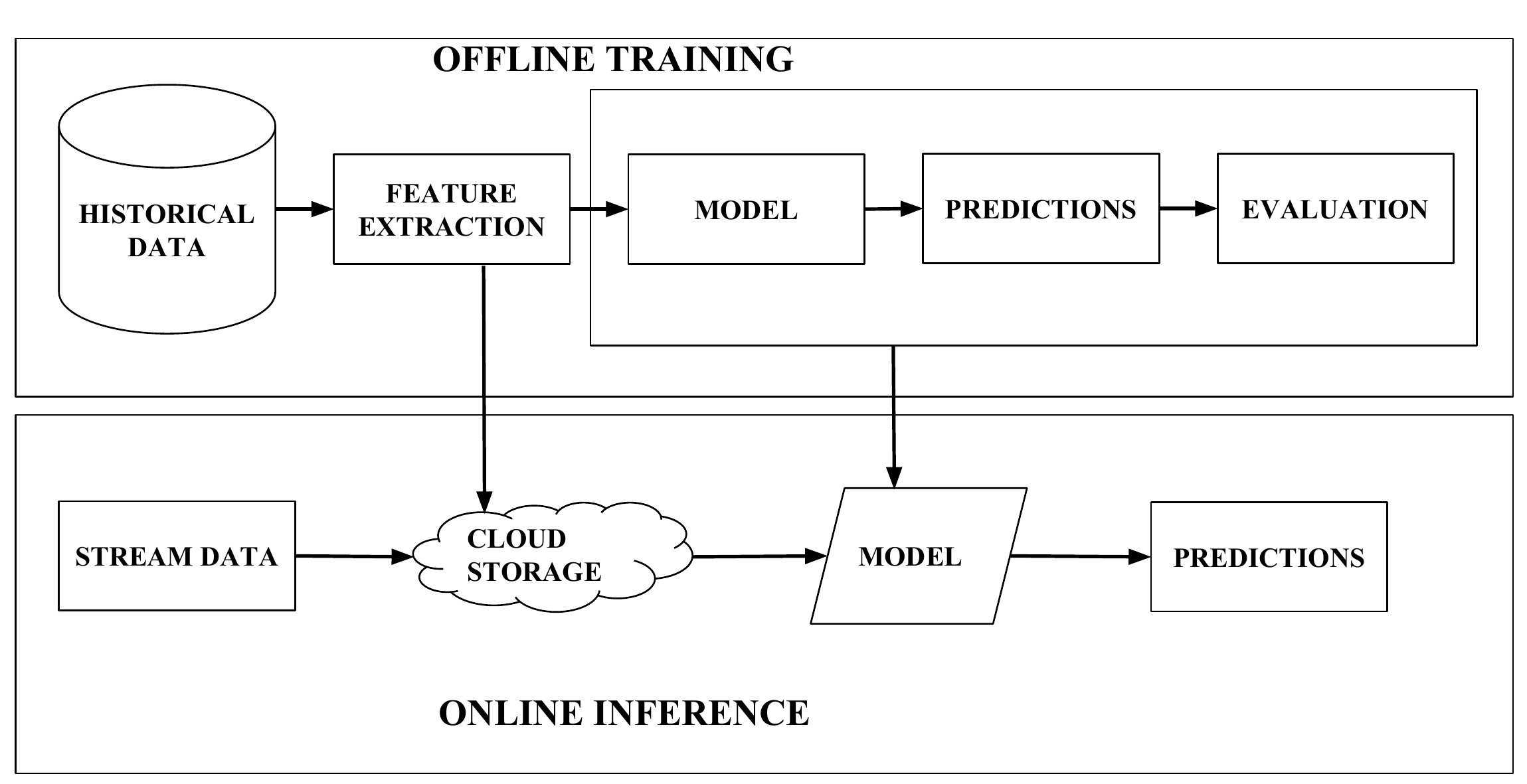}}}
    \caption{{Proposed framework of air pollution forecasting task}}%
    \label{fig:framework}
\end{figure}

Fig. \ref{fig:framework} demonstrates the framework of the proposed solution. It consists of two parts: Offline Training and Online Inference.

\subsection{{Offline Training}} The offline training involves the collection of data followed by data preprocessing which involves the extraction of the important features for air pollution from heterogeneous data and handling the missing values in the collected data for various location which is explained in the subsequent sections. The model is trained for the training data and then evaluated using the test set. The model obtained is deployed and is used for predictions in the online inference.

\begin{figure}%
    \centering
    {{\includegraphics[width=9cm, height=7cm]{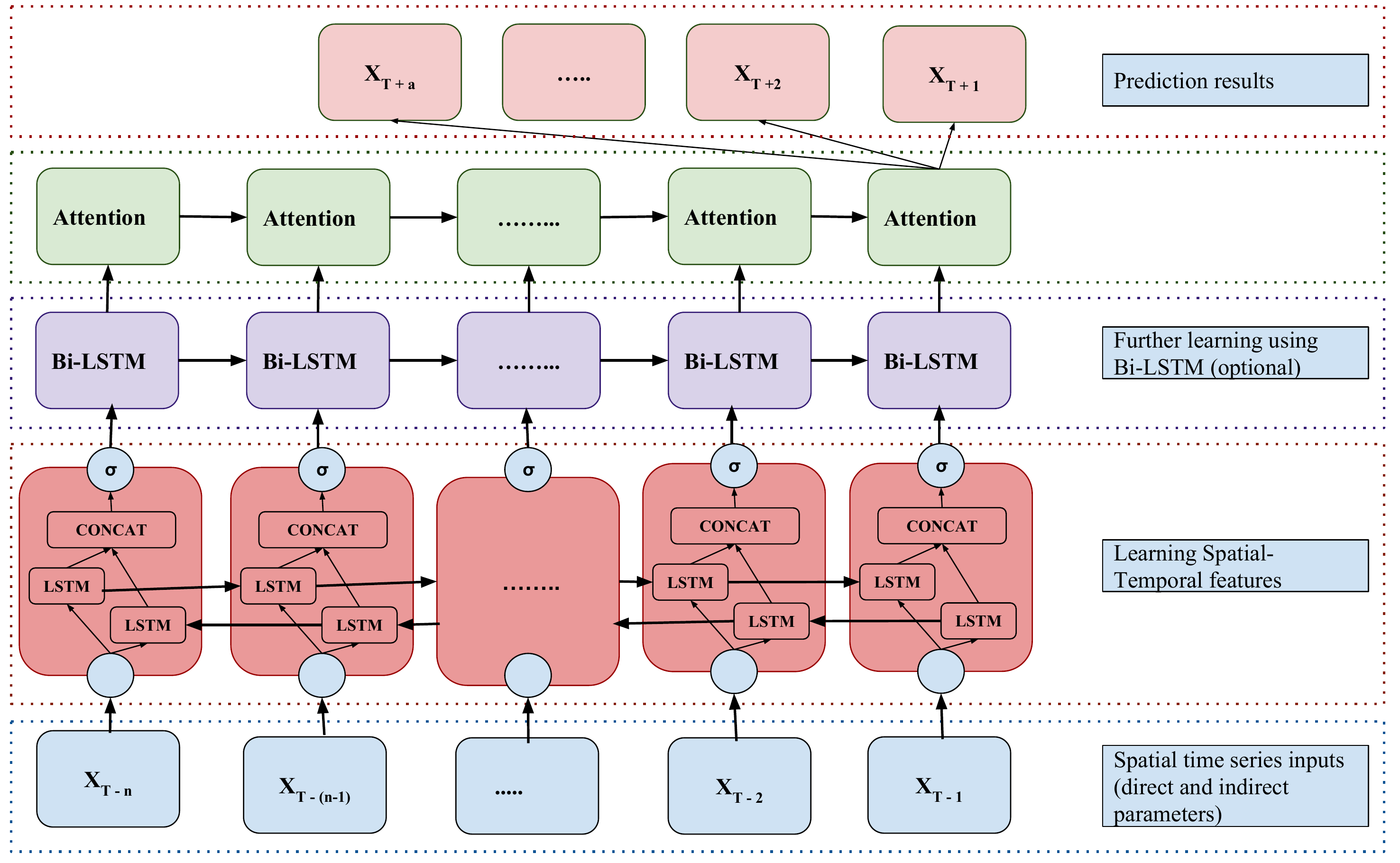}}}
    \caption{{Attention based BiLSTM model consists of a BiLSTM layer and the attention mechanism layer. $\{x_{T-n},x_{T-(n-1)},...x_{T-2},x_{T-1}\}$ represent the historical input data sequence (direct parameters- air pollutants, indirect parameters- meteorological data and time)to the BiLSTM layer and $\{x_{T+a},... x_{T+2}, x_{T+1}\}$ represent the pollution forecasted values. Multiple BiLSTM or LSTM layers are optional.}}
    \label{fig:finalModel}%
\end{figure}

\subsubsection{Data Collection}
The direct parameters consisting of all the air pollutants and indirect parameters which includes meteorological data and time, were collected for five air quality monitoring sites in Delhi from the \textbf{Central Pollution Control Board}\footnote{\href{http://cpcb.nic.in/}{http://cpcb.nic.in/}} database which has been widely used for model evaluation. CPCB has 703 operating stations in 307 cities/towns in 29 states and 6 Union Territories of the country, out of which 78 stations are in Delhi. The air pollutant data in this study includes  $SO_2$, $NO_2$, $PM_{2.5}$, $PM_{10}$, $CO$, $O_3$ and the meteorological data includes Temperature, Humidity, Wind Speed, Wind Direction, Barometric Pressure. \\ 
The data collected from the Central Pollution Control Board database had two main issues:
\begin{enumerate}
    \item The data collected from Central Pollution Control Board database contains missing values at random.
    \item The data may contain parameters that are not useful for the model for predictions and may act as noise.
\end{enumerate}

The complications faced due to the above mentioned issues in the data are addressed in the data preprocessing stage which is explained in the next subsection.

\subsubsection{Data Preprocessing}
\label{sec:datapreprocessing}
The data preprocessing step involves the selection of the important features for air quality forecasting and handling the missing values. The collected data contains missing values and some parameters that are not important for air quality prediction and sometimes acts as noise for the model.\\
i) Feature Extraction\\ 
The dataset is heterogeneous containing both direct and indirect factors. It is complicated to find the relationship these factors possess with air pollution. Although it is difficult to find the factors that impacts the air pollution, yet there are feature ranking methods that extract the most relevant features. We perform feature ranking using backward feature elimination, forward feature construction and Random Forests/Ensemble Trees to find the top ranked features for the proposed model. We construct a correlation matrix to remove the features having high values of correlation coefficients to avoid the redundancy of features. We obtain the temperature and humidity from the meteorological parameters and $SO_2$, $NO_2$, $PM_{2.5}$, $PM_{10}$, $CO$ and $O_3$ from the air pollution parameters after the feature ranking methods. We also obtain Hour of the day and months of the year as yet another important parameters using feature ranking methods.\\
ii) Data imputation \\
Since, the data consists of missing values at random (MAR) in the dataset for certain variables, it becomes a big challenge to handle them in order to obtain accurate forecasting of air pollution. Therefore, we impute the missing values by using the Multivariate Imputation by Chained Equations (MICE) \cite{mice}, also referred to as ``fully conditional specification" or ``sequential regression multiple imputation". We then apply normalization techniques to all the features and pollution targets so that the values lie in the range [0,1]. The reason behind normalization is to change the values of numeric columns in the dataset using a common scale, without distorting differences in the ranges of values or losing information.

\subsubsection{Model and Evaluation} 
We propose an attention based BiLSTM network as shown in Fig. \ref{fig:finalModel}. We utilize the trained model to forecast the pollutants for the next 24 hours on the test set which are used for the evaluation of different models.
The attention based BiLSTM consists of 4 modules:
\begin{enumerate}
\item Input Feature Module:
The input of the proposed model is the historical data sequence where each sequence is composed of pollution vectors represented as $P_f = \{p_1, p_2 ... p_n\}$, meteorological vectors represented as $M_f = \{m_1, m_2 ... m_n\}$ and time vectors represented as $T_f = \{t_1, t_2 ... t_n\}$. Finally, a sequence of combination of these vectors $S_f = \{P_f.M_f.T_f\}$ is sent to the BiLSTM module as inputs.
\item BiLSTM module:
This module consists of a single or multiple BiLSTM layers. For the first BiLSTM layer, the input is the historical data sequence represented as $S_f = \{P_f.M_f.T_f\}$ and the output is $h^1 = \{h^1_1, h^1_2...h^1_s\}$, where $h^1_t = [\overrightarrow{h^1_t}; \overleftarrow{h^1_t}]$ as described in subsection \ref{4b}.
\item Attention module:
This module consists of the attention layer which takes the outputs denoted as $h^n = \{h^n_1, h^n_2...h^n_s\}$, from the BiLSTM layer. This module assesses the importance of the representations of the encoding information $e_i$ and computes the weighted sum. The output of the attention function $f_{att}$ is normalized through a softmax function:
\begin{subequations}
\begin{align}
    z_{ji} = f_{att, j} (e_i) \\
    w_{ji} = \dfrac{exp(z_{ji})}{\sum_{k=1}^{E} exp(z_{jk)}}
\end{align}
\end{subequations}
\item Output module:
This module is the key module to get the network output. It consists of the fully connected layer which takes the $h^n = \{h^n_1, h^n_2...h^n_s\}$ as features. It generates the predicted values for the next 24 hours in the case of regression and the categorical values in case of classification of various pollutants.
\end{enumerate}
 Algorithm \ref{alg:the_alg} outlines the proposed adaptive method. By running the algorithm every week on the hourly updated data for each location, the errors of the online adaptive algorithm are minimized.
 
\begin{algorithm}
\KwIn {Data for each location $\{f_1, f_2,..f_{n-1}, f_n\}$; learning rate $\alpha$}
\textbf{Initialize} $F(x)$ = BiLSTM model with attention mechanism for $N$ pollutants  \\
\For{$t\gets1$ \KwTo $T$}{ 
 Receive instance: $x_t$ \\
Predict $\hat{y_t}$ for each pollutant for the next 24 hours \\
Recieve the true pollutant value $y_t$ \\
Suffer loss: $l_t(w_t)$ which is a convex loss function on both ${w_t}^Tx$ and $y_t$ \\
Update the prediction model $w_t$ to $w_{t+1}$
}
\caption{Algorithm for proposed adaptive method}
\label{alg:the_alg}
\end{algorithm}

\subsection{{Online Inference}} \label{3b}
The online inference as shown in Fig. \ref{fig:online} in the framework is composed of the following modules:
\begin{figure}[h!]
    \centering
    {{\includegraphics[width=8cm, height=6cm]{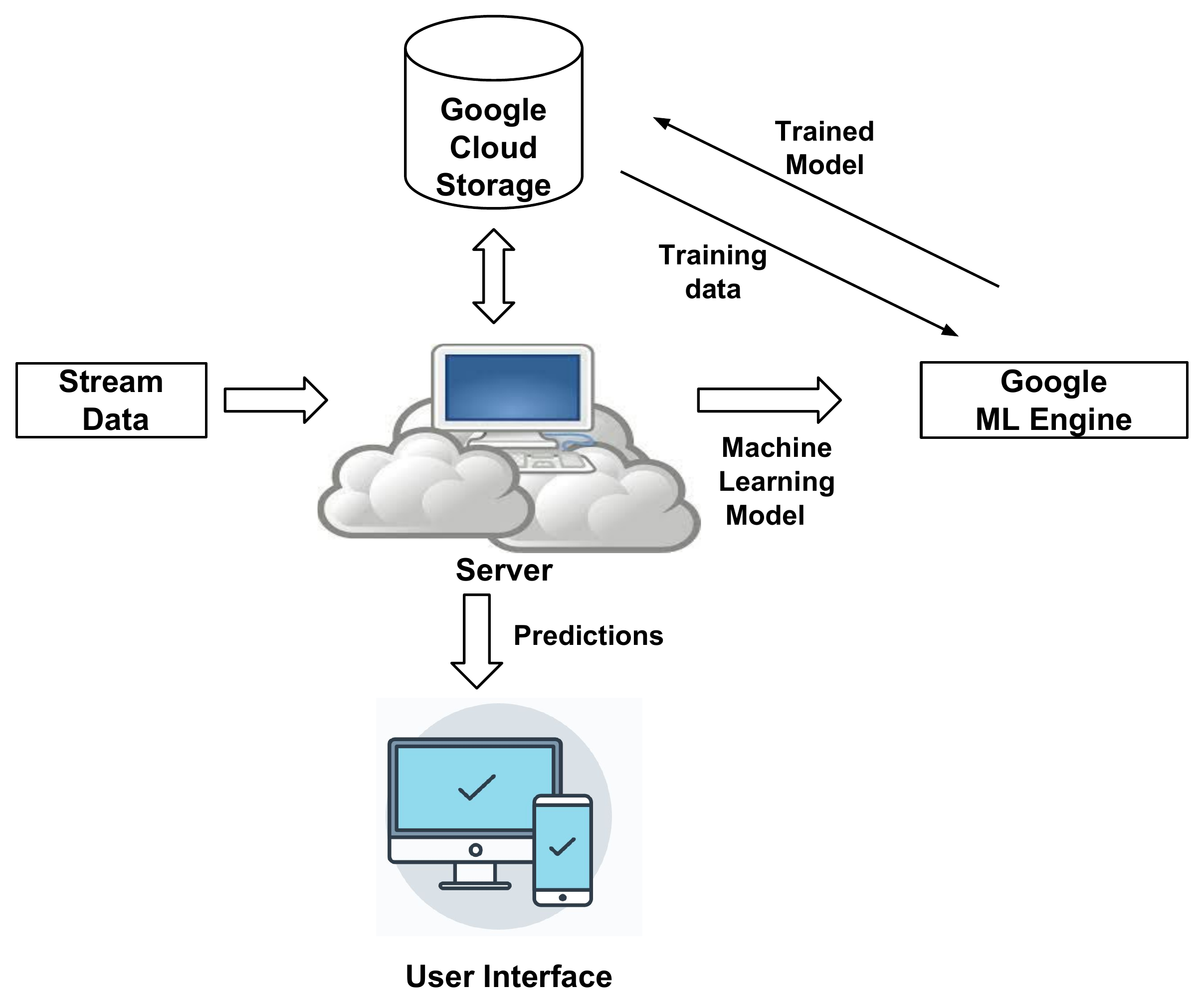}}}
    \caption{{Model for online inference}}
    \label{fig:online}%
\end{figure}
\subsubsection{Stream Data} The air quality data containing the air pollutant's concentration and meteorological data is collected real time for every hour and is updated on the server.
\subsubsection{Cloud Server and Cloud Storage} The server plays the prominent role of updating the real time data obtained every hour on the Cloud storage. After every week, the updated data along with the machine learning model is then passed by the server, where the training takes place. The trained model is then stored back on the Cloud Storage.
\subsubsection{User Interface} The stored model as part of the offline training process is then used to make the predictions every 4 hours for the next 24 hours and display it on the website \footnote{\url{http://bit.ly/cleanenv/}} for the user to observe the forecasted values. This would help the government and the people to take the necessary preventive measures to either control air pollution or to protect people from the harmful effects of air pollution.

\section{{Evaluation and Analysis}}
\label{sec:evaluation}
\subsection{Data Description} We utilize real time air quality monitoring dataset collected for 5 locations of Delhi from Central Pollution Control Board (CPCB) and Delhi Pollution Control Committee (DPCC) to evaluate the performance of various models. The dataset contains the Air Quality Data, time and Meteorological data. 
\subsubsection{Air Quality Data} The air pollutant variables in the air quality data are $SO_2$, $NO_2$, $PM_{2.5}$, $PM_{10}$, CO and $O_3$.
\subsubsection{Meteorological Data} The meteorological parameters include temperature, humidity, wind speed and barometric pressure.
\subsubsection{Time} Time includes hour of the day, seasons and month of the year. \\
All these variables of the past 24 hours are collected on hourly basis and extracted features from the collected dataset are used for evaluation of the models for predictions of the concentration of $PM_{2.5}$, $PM_{10}$ and $NO_2$ after every 4 hours for the next 24 hours. Concentrations of all the pollutants are reported in $\mu g/m^{3}$. 

\subsection{Implementation and Experimental Settings}
We split 80\% of the dataset for training and 20\% for test. We select 20\% of the training dataset as validation set and allow training to be early stopped according to the mean score error for the validations set. All values are then normalized in the range [0, 1]. The network is built using Keras with tensorflow as backend. For training, batch size is set as 64 and we utilize Adam \cite{adam} optimizer with learning rate of 0.001 for gradient descent optimization. Dropout \cite{dropout} is used as a regularization technique in all the models\footnote{Code will be released upon publication}.

\subsection{Results}
In this section, we evaluate the efficacy and efficiency of the proposed architecture against several baseline models and analyze the experimental results. We design prediction model to forecast the continuous and categorical values for different pollutants.
\subsubsection{Evaluation on Pollutants Values prediction task}
\begin{figure}[h!]
\centering
\begin{subfigure}[b]{0.49\textwidth}
\centering
   \includegraphics[width=\linewidth]{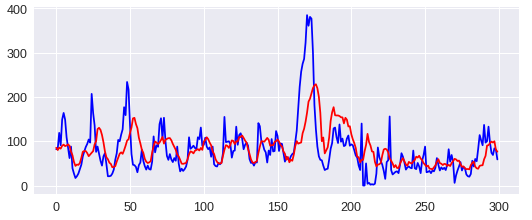}
   \caption{PM2.5}
   \label{} 
\end{subfigure}

\begin{subfigure}[b]{0.49\textwidth}
\centering
   \includegraphics[width=\linewidth]{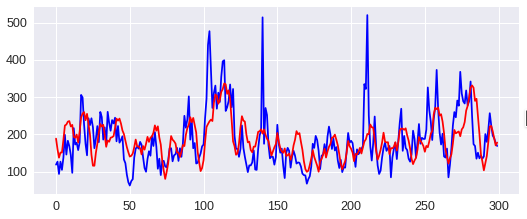}
   \label{}
   \caption{ PM10}
\end{subfigure}

\begin{subfigure}[b]{0.49\textwidth}
  \centering
   \includegraphics[width=\linewidth]{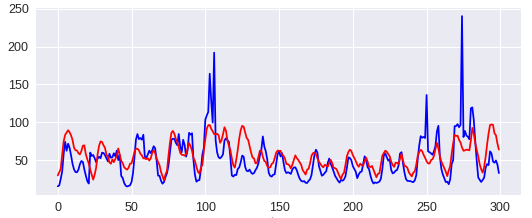}
   \caption{NO2}
   \label{}
\end{subfigure}
\caption{{Comparison between pollution estimates (red line) and actual measurements (blue line) for (a) $PM_{2.5}$ (b) $PM_{10}$ (c) $NO_2$.}
}
 \label{fig:reg_predictions}
\end{figure}

i) Experimental Design: It is a very important task to forecast the continuous values of the pollutants to circumvent the dangers caused by them and to take the necessary actions. For this task, we adopt regression loss as optimization function given as follows: \\
\begin{equation}
    L_{regression\_loss} = \sum_{X\epsilon{T}} (y(X) - \Hat{y}(X))^2 \\
\end{equation}
where $T$ denotes the set of all training instances with  $y(X)$ as ground truth values and $\Hat{y}(X) $ as predicted values. To make the predictions, we employ the following prediction models: LSTM (Long short term memory network), LSTM-A (Attention based Long Short term memory network), BiLSTM (BiDirectional Long Short Term memory network), BiLSTM-A (Attention based Bidirectional LSTM Network). In order to evaluate the performance of the various methods for regression, root mean square error (RMSE), and R-squared ($R^2$) are calculated.
\begin{subequations}
\begin{align}
RMSE = \sqrt{\dfrac{1}{n} \sum_{i=1}^{N} (\Hat{y_i} - y_i})^2 \\
R^2 = \dfrac{\sum_{i=1}^{N} (\Hat{y_i} - y_{avg} )^2} { \sum_{i=1}^{N} (y_i - y_{avg} )^2}
\end{align}
\end{subequations}
where $N$ denotes the number of instances in the test set, $\Hat{y}$ are the predicted values, $y$ denotes the actual values and $y_{avg}$ denotes the average of observations in the test set.

ii) Evaluation Results:
Table. \ref{tab1} shows the $R^2$ evaluation for different methods. It represents the evaluation for the next 4 hours. For all the instances, we use the historical values of the last 24 hours. 
In general, we observe a performance boost with BiLSTM-A in the predictions as compared to the other baseline models by $\sim 2 - 6\%$. Random Forest performs well for the forecasting of $PM_{2.5}$ values. The proposed model of BiLSTM-A outperforms in the prediction of all the other pollutants. Fig. \ref{fig:reg_predictions} shows the predictions results compared with the actual values for the BILSTM-A model for the next 4 hours. It shows that the proposed model achieves significant improvement in accuracy, especially in the scenarios of sudden change, which clearly demonstrates that the proposed method indeed benefits the temporal predictions.

\begin{table}[htbp]
\caption{Performance comparison of the proposed model with other baseline models for pollution values forecasting for future 4 hours on the basis of R-squared values and Root mean square error values. Highlighted values indicates the best performance.}
\begin{center}
\begin{tabular}{|c|c|c|c|}
\hline
\textbf{Model} & \textbf{\textit{Pollutants}}& \textbf{\textit{R-square}}& \textbf{\textit{RMSE}} \\
\hline
                &  $PM_{2.5}$ & \textbf{0.35} & \textbf{40.69} \\ 
      Random Forest & $NO_2$ & 0.406 & 21.12\\ 
                 & $PM_{10}$ & 0.425 &  98.32 \\
\hline
                &  $PM_{2.5}$ & 0.31 & 41.96 \\ 
      LSTM & $NO_2$ & 0.383 & 21.52\\ 
                 & $PM_{10}$ & 0.445 &  96.58 \\
\hline
  &  $PM_{2.5}$ & 0.29 & 42.52 \\ 
      LSTM-A & $NO_2$ & 0.387 & 21.44\\ 
                 & $PM_{10}$ & 0.446 &  96.49\\
\hline
  &  $PM_{2.5}$ & 0.30 & 42.07 \\ 
      BILSTM & $NO_2$ & 0.385 & 21.47\\ 
                 & $PM_{10}$ & 0.442 &  96.77 \\
\hline
  &  $PM_{2.5}$ & 0.310 & 41.97 \\ 
      BILSTM-A & $NO_2$ & \textbf{0.417} & \textbf{21.08}\\ 
                 & $PM_{10}$ & \textbf{0.454} &  \textbf{96.22} \\
\hline
\end{tabular}
\label{tab1}
\end{center}
\end{table}

\subsubsection{Evaluation on Pollutants Levels prediction task}
i) Experimental Design: Most of the previous literature work \cite{dong2009,Sun2017,zheng15,zhu18,corani2016} is based on prediction of continuous values. We emphasize on the effectiveness of the prediction for various pollutant levels. For this task, the concentration levels are divided into different threshold values by considering the critical values for each of them. The threshold values are shown in Tab. \ref{tab:classes}
\begin{table}[htbp]
\begin{center}
\caption{\textbf{Pollution Concentration Levels}}\label{tab:classes}
\begin{tabular}{|c|c|c|}
\hline
    \textbf{Pollutant} &\textbf{Range}&\textbf{Class}  \\ 
\hline
                &  0 - 60 & 0\\ 
      $PM_{2.5}$ &  60 - 150 & 1 \\ 
                 & 150+ & 2 \\
\hline
             & 0 - 50 & 0\\ 
    $NO_{2}$ & 50+   & 1 \\ 
\hline
          & 0 - 100 & 0\\ 
 $PM_{10}$ & 100 - 250 & 1\\ 
           & 250+ & 2 \\ 
\hline
\end{tabular}
\end{center}
\end{table}

For this task, we adopt the optimization function is chosen as the classification loss given as follows: \\
\begin{equation}
    L_{classification\_loss} = - \sum_{c=1}^{M} y_{o,c} \log({p_{o,c}})
\end{equation}
where $M$ denotes the number of classes, $y$ is a  binary indicator (0 or 1) of whether class label $c$ is the correct classification for observation $o$ and $p$ is the model's predicted probability that observation $o$ belongs to class $c$. To make the predictions we employ the following prediction models: LSTM (Long short term memory network), LSTM-A (Attention based Long Short term memory network), BiLSTM (BiDirectional Long Short Term memory network), BiLSTM-A (Attention based Bidirectional LSTM Network). We categorize the output forecasted by the regressor models and compare them with the output of the classifier model from CBiLSTM-A, which is a BiDirectional LSTM used for classification. In order to evaluate the performance of the various methods for classification, we calculate the accuracy, precision and recall rates.
\begin{equation}
Accuracy = \dfrac{|\{X \epsilon TestSet | \Hat{Y}(X) = y(X)\}|} {N}    
\end{equation}

where $N$ is the number of instances in the test set, $\Hat{Y}$ are the predicted values and $y$ are the actual values.\\
\begin{equation}
Precision = \dfrac{{true\,positives}} {true\,positives + false\,negatives}    
\end{equation}
Precision is defined as the number of true positives divided by the number of true positives plus the number of false positives. False positives are cases the model incorrectly labels as positive that are actually negative, or in our example, individuals the model classifies as terrorists that are not. \\
\begin{equation}
Recall = \dfrac{{true\,positives}} {true\,positives + false\,positives}    
\end{equation}
 Recall is the number of true positives divided by the number of true positives plus the number of false negatives. True positives are data point classified as positive by the model that actually are positive (meaning they are correct), and false negatives are data points the model identifies as negative that actually are positive (incorrect). \\
ii) Evaluation Results:
Table \ref{accuracy} shows the evaluation of  pollution level forecasting by comparing accuracy, precision, recall and F1 Score of all the models. It represents the evaluation for the next 4 hours. For all the instances for the classification task we use the historical values of last 24 hours.

\begin{table}[htbp]
\caption{Performance comparison of the proposed model with other baseline models for pollution levels forecasting for future 4 hours on the basis of Accuracy, average precision and average recall. Higher values of accuracy, precision and recall indicates the better performance of the model. Highlighted values indicates the best performance.}
\begin{center}
\begin{tabular}{|c|c|c|c|c|}
\hline
\textbf{Model} & \textbf{\textit{Pollutants}}& \textbf{\textit{Accuracy}}& \textbf{\textit{Precision}} & \textbf{\textit{Recall}}  \\
\hline
                &  $PM_{2.5}$ & 67.68 & 56.15 & 52.27\\ 
      LSTM & $NO_2$ & 76.85 & 76.29 & 75.2\\ 
                 & $PM_{10}$ & 68.34 & 71.11 & 56.31 \\
\hline
  &  $PM_{2.5}$ & 67.24 & 56.46 & 52.56\\ 
      LSTM-A & $NO_2$ & 76.85 & 76.15 & 75.65\\ 
                 & $PM_{10}$ & 68.71 & 70.21 & 57.89 \\
\hline
  &  $PM_{2.5}$ & 67.96 & 58.35 & 53.12\\ 
      BILSTM & $NO_2$ & 77.32 & 76.75 & 75.86\\ 
                 & $PM_{10}$ & \textbf{68.87} & \textbf{70.25} &  58.36 \\
\hline
  &  $PM_{2.5}$ & 67.96 & 55.71 & 52.55\\ 
      BILSTM-A & $NO_2$ & 77.66 & 77.1 & \textbf{76.26}\\ 
                 & $PM_{10}$ & 68.21 & 69.21 & 57.73 \\
\hline
  &  $PM_{2.5}$ & \textbf{70.68} & \textbf{61.06} & \textbf{55.8} \\ 
      CBILSTM-A & $NO_2$ & \textbf{77.88} & \textbf{77.56} & 76.14\\ 
                 & $PM_{10}$ & 67.45 & 68.23 & \textbf{58.52} \\
\hline
\end{tabular}
\label{accuracy}
\end{center}
\end{table}

 It shows that BiLSTM-A shows slight in accuracy compared to the other models used in the air pollution prediction task.
BiLSTM with attention (BiLSTM-A) is preferred over BiL-
STM because when Attention mechanism is applied on top of
BiLSTM, it captures periods and makes BiLSTM more robust
to random missing values.
\subsection{Trends in Pollution}
Delhi is located at 28.61\textdegree N 77.23\textdegree E, and lies in Northern India. The area represents high seasonal variation. Delhi is covered by the Great Indian desert (Thar desert) of Rajasthan on its west and central hot plains in its south part. The north and east boundaries are covered with cool hilly regions. Thus, Delhi is located in the  subtropical belt with extremely scorching summers, moderate rainfall, and chilling winters.
\begin{figure}[htbp]
\centering
\begin{subfigure}[b]{0.49\textwidth}
  \includegraphics[width=\linewidth]{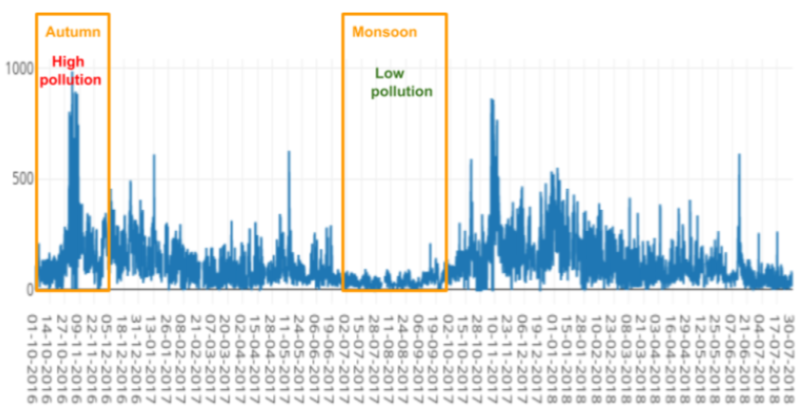}
   \caption{PM2.5}
  \label{} 
\end{subfigure}

\begin{subfigure}[b]{0.49\textwidth}
  \includegraphics[width=\linewidth]{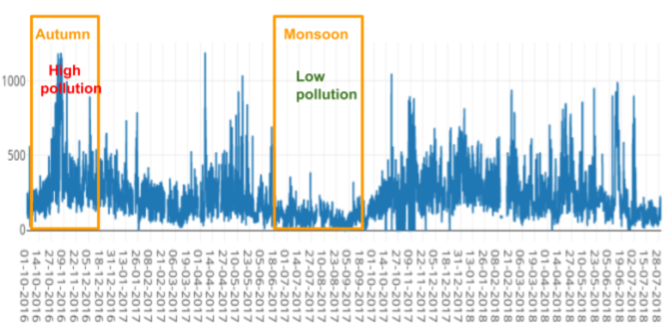}
   \caption{PM10}
  \label{}
\end{subfigure}
\begin{subfigure}[b]{0.49\textwidth}
  \includegraphics[width=\linewidth]{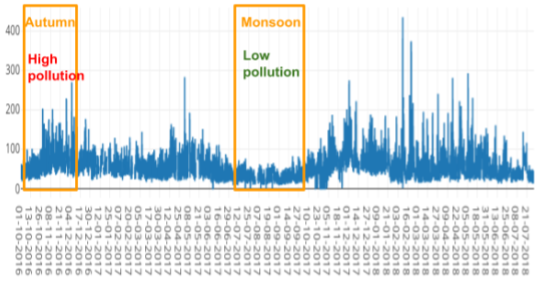}
   \caption{NO2}
  \label{}
\end{subfigure}
\caption{{Seasonal variation in concentration of Mandir Marg, Delhi for (a) $PM_{2.5}$ (b) $PM_{10}$ (c) $NO_2$.} X-axis represent date and Y-axis represent concentration of pollutant.}
 \label{fig:seasonal}
\end{figure}

As the concentrations of each of the air pollutants is directly related to seasonal variation of the atmosphere, it becomes important to study the break-up of 12 months. Winter season includes December and January months, spring season includes February and March months, summer season includes April, May and June months, monsoon season includes July, August and September months while autumn includes October and November months.

Fig. \ref{fig:seasonal} shows the concentration trends for Mandir Marg location in Delhi, in the five seasons (winter, spring,  summer, monsoon and autumn), from October 2016 to July 2018. The concentration of Mandir Marg area was comparatively worst in winter and best in monsoon. The following sequence in the decreasing order was observed in the concentration of $PM_{10}$, $PM_{2.5}$ and $NO_2$: winter $>$ autumn $>$ spring $>$ summer $>$ monsoon. Decreasing order of concentration implies air quality going from worst to better, it means here that winter has worst air quality and monsoon has best.

High concentration is observed after 4pm due to vehicular emission.
The concentration of $NO_2$ is highest during  8 am - 10:30 am and 4 pm - 7 pm  due to the vehicular emission, from the burning of fuel, from emissions from cars, trucks, buses etc.
\subsection{Real Time Learning System}
\begin{figure}[htbp]
    \centering
     {{\includegraphics[width=9cm, height=3cm]{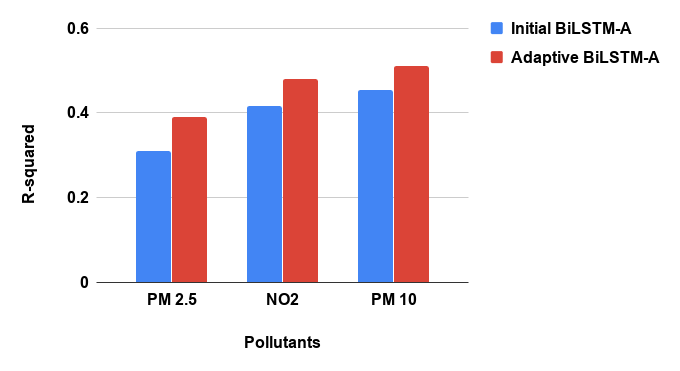}}}
    \caption{{Performance of  Initial BiLSTM-A model with initial data compared to  Adaptive BiLSTM-A model
}}
    \label{fig:continuos}%
\end{figure}

The pollution forecasting and source prediction methods discovered in this study can help governments and people take necessary sections. We train the initial model used in the evaluation on the initially collected data. Due to the progressive temporal changes in the concentration of the pollutants, it is necessary to continuously update the collected data and the model, thus resulting into an adaptive model. We update the training data by collecting it from the Central Pollution Control Board every hour as explained in subsection \ref{3b}.
In this section, we compare the performance of the model on the initial collected data to the performance of the model which is updated every week on the real time hourly updated data after a month of the collection of the initial data. BiLSTM-A model is used for evaluation of both the models with the updated data in Fig. \ref{fig:continuos}.
 \\
Fig. \ref{fig:continuos} shows substantial improvement of the results after the real time update which is indicative of the relevance of continuous updation of the model with the real time incoming data.
\section{{Web Application}}
\label{sec:webapp}
The Air quality index is an index for reporting daily air quality. It tells you how clean or polluted your air is, and what associated health effects might be a concern for you. The AQI focuses on health effects you may experience within a few hours or days after breathing polluted air. The higher the AQI value, the greater the level of air pollution and the greater the health concern. For example, an AQI value of 50 represents good air quality with little potential to affect public health, while an AQI value over 300 represents hazardous air quality. Our website works for five locations of Delhi.

\begin{figure}[h!]
\centering
\begin{subfigure}[b]{0.49\textwidth}
   \includegraphics[width=\linewidth]{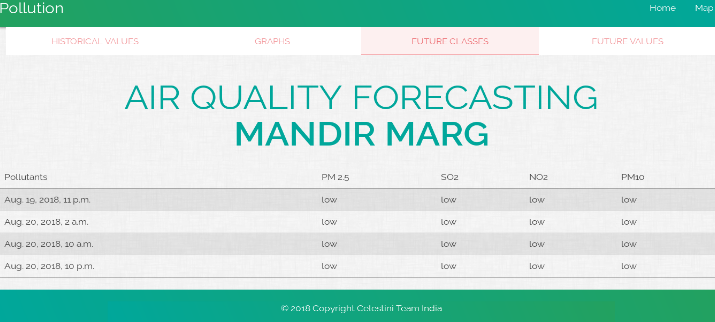}
  \caption{}
   \label{}
\end{subfigure}

\begin{subfigure}[b]{0.49\textwidth}
   \includegraphics[width=\linewidth]{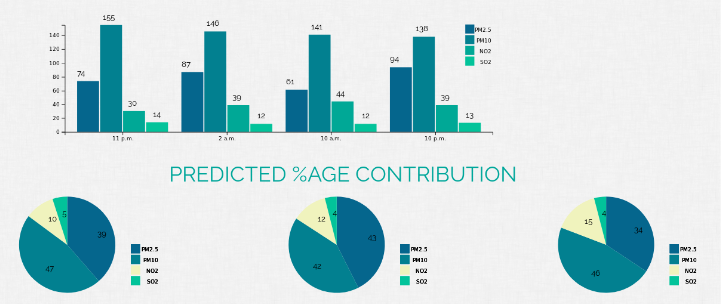}
   \caption{}
   \label{}
\end{subfigure}
\caption{
Two functionalities of our website are
 (a) Pollution level forecasting (b) Forecasting of exact concentration of pollutants.
}                             
 \label{fig:website}
\end{figure}

Sample  screen  shots  of  our  web  application for Mandir Marg location of Delhi are  shown
in  Fig. \ref{fig:website}.  Our web application comprises of four main functionalities: 

\begin{enumerate}
    \item Pollution percentage contribution and source determination: The grouped bar charts and pie charts for different pollutants including PM2.5, PM10, NO2 are displayed 
    The graphs are displayed at an interval of 4 hours of a day to help you observe different trends during the day to keep you safe for the next.
    \item Historical concentration of pollutants: 
Monitor pollution and air quality in Delhi using our air quality data. We display the Air Quality index for the past 24 hours for various pollutants
The values are displayed to display the trends in the past 24 hours to help you make informed decisions about what activities you do outside, when and for how much time.
    \item Pollution level forecasting: We predict the pollution level (eg. low, moderate and high pollution) for different air pollutants including nitrogen dioxide ($NO_2$) and particulate matter ($PM_{2.5}$ and $PM_{10}$)
    We classify the pollution level for each pollutant to 2-3 classes ( 3 classes for $PM_{2.5}$ and $PM_{10}$, 2 classes for  $NO_2$) and forecast the pollution level for each pollutant for the next 24 hours.
    \item Forecasting of exact concentration of pollutants: We display grouped bar charts and pie charts for predicted concentrations of different pollutants including PM2.5, PM10, NO2 
    The graphs are displayed at an interval of 4 hours of a day and helps you to observe different trends during the day.
\end{enumerate}
    
\section{{Conclusion}} 
\label{sec:conclusion}
Based on the historical and real-time ambient air quality and meteorological data of 5 monitoring stations in Delhi, we inferred the real-time and fine-grained ambient air quality information. We build a novel end-to-end system to predict the air quality of next 24 hours by predicting the concentration and the level (low, moderate, high) of different air pollutants including nitrogen dioxide ($NO_2$), particulate matter ($PM_{2.5}$ and $PM_{10}$) for Delhi. Extensive experiments on air pollution data for 5 locations in Delhi, helped evaluating the proposed approach. The results showed the performance boost with the proposed method over other well known methods for classification and regression models. The video\footnote{\url{http://bit.ly/pollution_video}} describes our work in brief.
In future work, we intend to explore more powerful modeling techniques along with the traffic density data, as a way to model the traffic density of the monitored location to get better results.

\section*{Acknowledgments}\label{sec:Acknowledgments}
We would like to thank Dr Aakanksha Chowdhery, Google Brain for the inspiring  guidance and constant encouragement during the course of this work. We acknowledge the Marconi Society, IIT Delhi and Celestini Program India to support the project.
We would also like to thank Central pollution control board for the dataset which has played a very important role towards the completion of this work.
\bibliographystyle{plain}
\bibliography{references}

\end{document}